# An Autonomous Adaptive Learning System for Robotic Grasping in the Circular Factory

Yitian Shi[0009-0007-3713-7739], Rosa Wolf[0000-0003-3544-8203], Rania Rayyes[0000-0001-7733-4509]

Karlsruhe Institute of Technology (KIT), Gotthard-Franz Straße 8, Karlsruhe, 76131, Germany
yitian.shi@kit.com

**Abstract.** Linear manufacturing typically operates in highly structured settings with standardized objects, allowing robots to execute pre-programmed motions based on precise models. In contrast, circular manufacturing involves dynamic and less controlled environments, where product components can vary due to appearance and physical changes across different product generations. In this context, uncertainty arises from both the unknown or partially observed states of components (e.g., damaged surfaces, altered shapes) and the unstructured disassembly environment, often characterized by heavy clutter, occlusions, and limited sensing viewpoints. This demands a flexible and adaptable robotic grasping system that can cope with unexpected object changes. To address these challenges, this paper proposes an autonomous grasp learning framework designed for highly uncertain and variable conditions. The framework explicitly distinguishes between uncertainty due to noisy or incomplete sensory observations (aleatoric uncertainty) and uncertainty due to limited knowledge and distribution shift (epistemic uncertainty). Moreover, the framework employs an active perception strategy to reduce visual ambiguity: semantic and spatial scene understanding is achieved using vision language models (VLMs), while view planning enables the robot to adjust its viewpoint to improve observability. In addition, the approach leverages large-scale human demonstration data to learn interaction patterns, which are then transferred to robotic end-effectors, enabling robust manipulation in these variable settings. In summary, our system can: (1) adapt grasping strategies to uncertain objects, (2) actively search for partially or fully occluded objects in cluttered scenes, and (3) continuously adapt its policy using human demonstrations. Quantitative validation, including the utility of active vision on disassembled components in practice, is expected to form an important part of future work.

**Keywords:** robotic grasping, intralogistics, material handling, circular factory, autonomous and intelligent system.

## 1 Introduction

Robotic grasping has achieved significant progress in structured industrial settings, where objects are standardized, process conditions are tightly controlled, and perception pipelines operate with limited ambiguity. Such assumptions are often valid in linear manufacturing, in which the robot handles known objects, at known poses, under stable sensing and lighting conditions. In circular manufacturing, these assumptions break down, where robots deployed for disassembly, remanufacturing, sorting, and component recovery must handle products that are aged, damaged, occluded, partially disassembled, or mixed across product generations (as illustrated in Fig. 1.). As a result, both the geometry of the object and the context in which it appears can differ substantially from what was observed during algorithm training for detection or grasp generation.

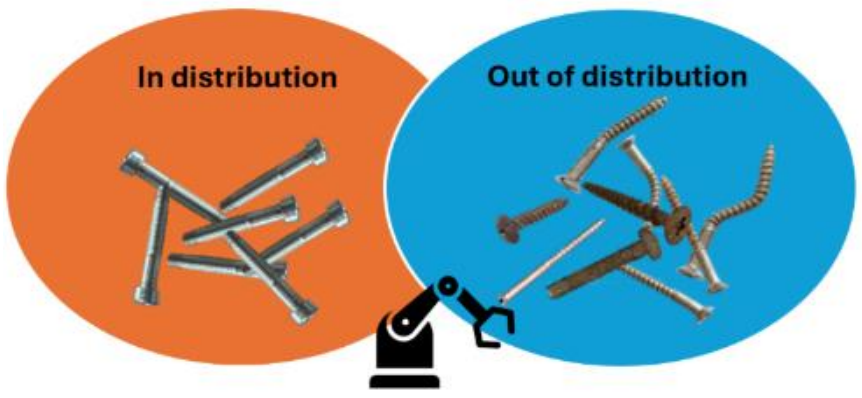

**Fig. 1.** While linear production involves uniform objects with known geometries that are easily localized via pose matching (i.e., objects are typically "in distribution" (ID)), circular factories introduce significant distribution shifts. The high variance and unknown states of recycled materials require advanced, uncertainty-aware manipulation strategies to reliably grasp "out-of-distribution" (OOD) objects in cluttered environments.

Subsequently, this mismatch severely limits traditional analytic grasping approaches, which typically rely on precise geometric models and assume perfect state estimation. To overcome these limitations, learning-based robotic grasp generation [1] has emerged as the primary alternative. However, to operate reliably in unstructured environments, these learning-based models must explicitly account for two distinct but interacting sources of uncertainty [2]: The first is aleatoric uncertainty, which arises from noisy sensing, incomplete view, and reflective or damaged surfaces; The second is epistemic uncertainty, which occurs when the robot encounters novel object instances, distribution shifts between training and deployment, or the ambiguities that cannot be fully resolved from a single observation. Therefore, a robust circular manufacturing system must not only detect grasp affordances under these conditions, but actively leverage these uncertainty estimates to know when the scene should be re-observed, when the policy should explore to seek informative interactions, and how prior knowledge can be transferred from humans to robots.

From this perspective, recent work has addressed individual aspects of this challenge. Uncertainty-driven online grasp learning [3] has shown that grasp exploration can be guided by uncertainty estimates to improve data efficiency and adaptation during deployment. Moreover, probabilistic contact-grasp reasoning using evidential learning and von Mises-Fisher (vMF) directional representations has enabled more reliable 6-DoF grasping in noisy clutter by modeling both epistemic and aleatoric uncertainty [4]. To mitigate the aleatoric uncertainty during online inference, vision-language-informed active view planning has demonstrated the importance of semantic and spatial scene understanding for discovering target objects under clutter and invisibility [5].

This paper proposes CIRCA-Grasp, a unified framework that combines these complementary ideas for robotic grasping in circular manufacturing. The key insight is that reliable behavior under the circular factory environments requires more than a single grasp predictor. Instead, the robot must reason over multiple layers of uncertainty, actively choose better observations, exploit prior human interaction knowledge, and continue adapting through online experience.

In summary, our contributions are threefold:

- We propose CIRCA-Grasp, a closed-loop robotic grasp learning framework that integrates digital assets, human demonstration priors, and real-world robot experience for manipulation in highly uncertain circular-manufacturing environments.
- CIRCA-Grasp couples vMF-Contact for probabilistic uncertainty-aware grasp prediction with VISO-Grasp for active view planning, allowing the robot to search for, observe, and grasp partially visible or occluded objects in cluttered scenes.
- We demonstrate the seamless integration and connections between CIRCA-Grasp and the knowledge-based intralogistics system proposed by [14].

In general, CIRCA-Grasp is presented as a conceptual integration framework in this paper, whose primary contribution lies in defining the interfaces, shared representations, and uncertainty-aware decision logic required to integrate grasp learning, active perception, human-derived priors, and knowledge-based intralogistics.

## 2 Related Work

### 2.1 Uncertainty-aware robotic grasping

Uncertainty estimation [2] has become increasingly important for robotic manipulation in realistic settings. Prior work on online grasp learning has shown that epistemic uncertainty can guide exploration toward informative samples, allowing robots to improve grasp policies more efficiently during interaction [3]. Such approaches are particularly relevant in deployment settings where labels are sparse and failures must be converted into useful learning signals. However, exploration alone is insufficient if the robot cannot distinguish a lack of knowledge from intrinsically ambiguous observations.

Probabilistic contact-based grasp formulations offer a natural way to reason about uncertainty in grasp geometry. vMF-Contact [4] introduced evidential learning for probabilistic contact-grasp estimation [6], representing contact directions with von Mises-Fisher distributions and modeling both aleatoric and epistemic uncertainty in noisy clutter. This representation is critical for robotic grasping in the context of circular manufacturing, where surface degradation, damage, or occlusion often makes deterministic grasp prediction brittle.

### 2.2 Active perception for occluded target grasping

In heavily cluttered scenes, single or static viewpoints [6, 7] for grasp prediction often fail because the robot lacks sufficient evidence to localize the target. While conventional active perception and Next-Best-View (NBV) strategies typically rely on geometric information gain or volumetric frontier exploration to uncover hidden areas [8], they often lack the semantic context required to efficiently search for specific, fully occluded objects. Bridging this gap, VISO-Grasp [5] integrates vision-language understanding with spatial object-centric reasoning to infer the locations of invisible targets based on their relationships with visible anchors, directly guiding 6-DoF active view planning. Compared to purely geometry-driven or reinforcement learning-based active vision methods [8, 9], VISO-Grasp's semantic scene understanding is particularly advantageous for circular manufacturing. Rather than relying solely on concrete geometric priors, it enables the system to proactively locate and grasp targets defined by component-level relational descriptions, amidst the unpredictable variance of recycled materials.

### 2.3 Human demonstration priors for grasp synthesis

Large-scale human manipulation data provides rich, natural priors. For instance, contact patterns, hand shapes, and functional affordances are difficult to capture through robot-centric exploration alone. Recent advancements have successfully leveraged these priors for generative grasp synthesis, demonstrating that taxonomy awareness and hand-object interaction modeling can significantly constrain the search space for stable grasps [10, 11]. In the context of circular manufacturing, these human-guided priors are exceptionally valuable, since humans intuitively adapt their grasping strategies to handle object wear, partial damage, and varied task contexts. Transferring these robust interaction regularities to robotic end-effectors offers a promising pathway to generalize across the high geometric variance and unknown states of recycled components.

However, a significant morphological gap exists, since human demonstrations inherently assume a highly articulated, multi-fingered hand, whereas factory floors predominantly rely on robust parallel-jaw grippers. Transferring complex human interaction regularities to these morphologically disparate end-effectors remains a major challenge, which is simplified by only demonstrating pinch grasps in much of the literature [12,13]. To address this discrepancy, CIRCA-Grasp leverages HOGraspFlow [10], a taxonomy-aware grasp retargeting framework to seamlessly translate rich human demonstration priors into actionable, generalized grasps for parallel-jaw end-effectors.

## 3 CIRCA-Grasp

### 3.1 Framework overview

As shown in Fig. 2., CIRCA-Grasp is a closed-loop grasp learning framework that combines digital assets, physical assets, and deployment-time active perception for robotic grasping in circular manufacturing. The framework is built around a shared grasp learning database that aggregates three complementary data sources:

- Synthetic grasp data generated from CAD models and simulated clutters;
- Transferred human interaction priors from demonstration data;
- Real-world robot experiences collected during deployment.

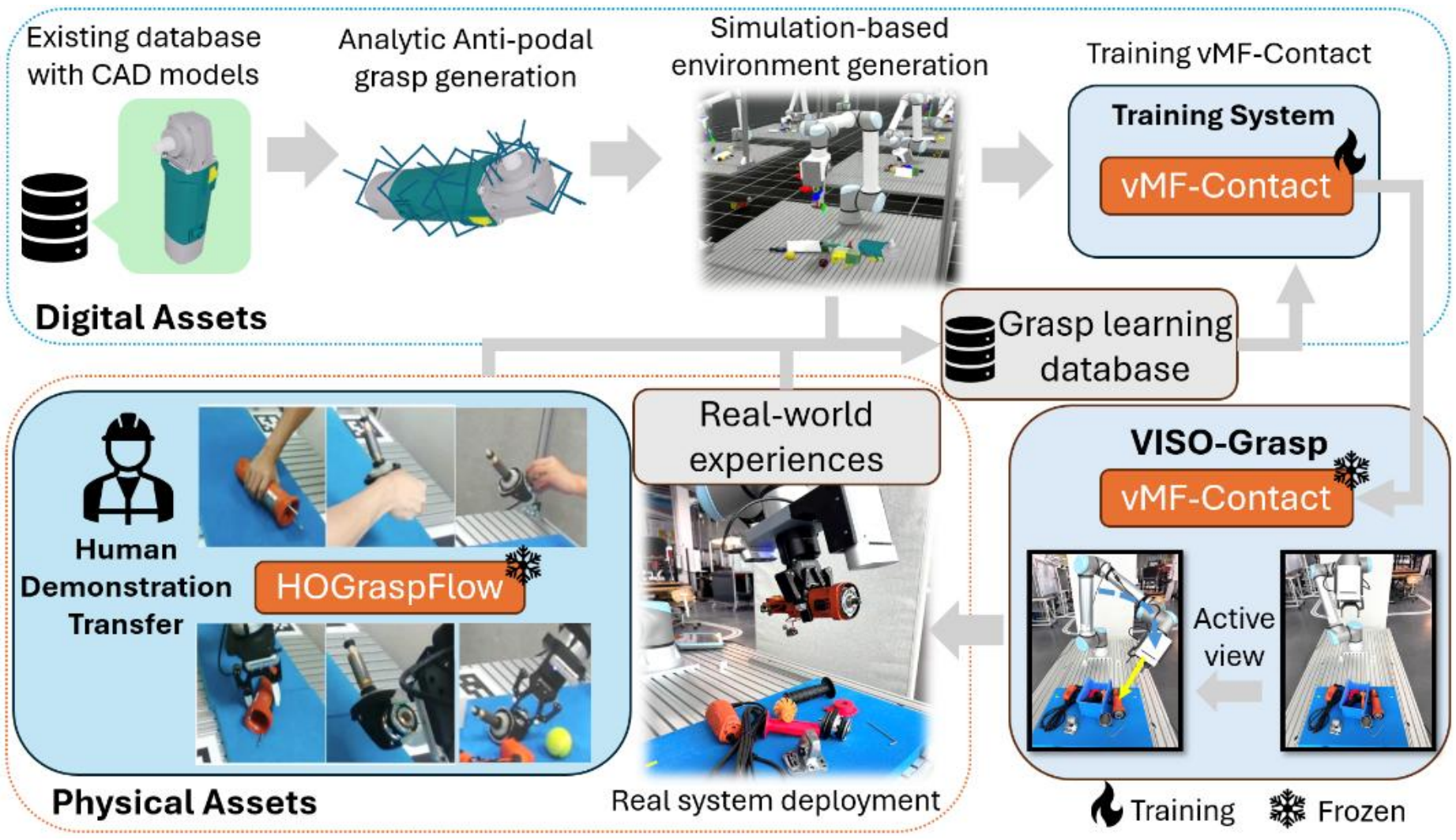

**Fig. 2.** Overview of the proposed CIRCA-Grasp framework. In the digital asset pipeline, existing CAD models are used to generate analytic antipodal grasps and simulation-based training environments, which support pretraining of the probabilistic grasp learner vMF-Contact [4]. In the physical asset pipeline, large-scale human demonstrations are transferred through HOGraspFlow [10] to provide hand-object interaction priors, while real-world robot executions continuously generate grasp experiences for deployment and adaptation. During inference, VISO-Grasp [5] integrates the uncertainty-aware grasp model with active view planning to search for and grasp partially visible or occluded objects in clutter. All simulated and real interactions are stored in a grasp learning database, enabling continual improvement of the grasp policy across perception, planning, and execution.

The three data sources are integrated through a shared grasp-learning database. Each database record contains a scene observation and corresponding 6-DoF grasp pose(s). When available, the record additionally stores the selected viewpoint, visibility estimate, aleatoric and epistemic uncertainty, and model version. Human demonstration records contain the demonstration identifier and HOGraspFlow output. Real-world records contain the executed robot pose, viewpoint history, sensor-derived outcome, and verification status. This common schema allows all three sources to be retrieved and sampled through the same training interface without discarding their provenance.

### 3.2 Learning from digital assets

The upper branch of Fig. 2. corresponds to the digital asset pipeline. As in linear manufacturing, the digital asset pipeline begins with an existing database of CAD models, representing target objects or component families. Although the deployed objects in circular manufacturing may differ from their nominal models due to wear, deformation, or product variation, CAD data still provides a strong geometric prior for initializing grasp learning.

From these CAD models, we first generate analytic antipodal grasps to obtain physically plausible contact configurations following [1, 4]. We then place the objects into simulation-based environments with varying clutter, poses, occlusions, and sensor conditions to approximate the uncertainty encountered in real disassembly scenes. Notably, this is done using the MetaIsaacGrasp framework [4], which is developed upon Isaac Lab [15] for efficient parallel scene generation. After that, the grasps that may collide with the environment will be filtered out through the collision check. These scene-level non-collision grasps serve as structured supervision for grasp representation learning, which are stored in the grasp learning database.

Using this synthetic grasp database, we train the vMF-Contact network as the initialization of our core grasp learning agent. Compared to general 6-DoF grasp learning frameworks [1, 6], it predicts probabilistic contact-grasp hypotheses rather than deterministic grasp poses, allowing the model to explicitly represent ambiguity in local geometry and visibility and disentangle the source of uncertainties.

### 3.3 Incorporating human demonstration priors

The lower-left branch of Fig. 2. corresponds to human demonstration transfer through HOGraspFlow, a learning-based hand-gripper retargeting framework. Human demonstrations complement synthetic data by providing functional contact and approach priors under object variation. To bridge the gap between in-the-wild human demonstrations and robotic end-effectors, HOGraspFlow learns these interaction priors from human demonstrations and transfers them to robotic grasps. In this way, given a grasp demonstration, HOGraspFlow injects task- and interaction-level knowledge into the framework, improving robustness beyond geometry-driven learning.

To make the retargeted human priors commensurable with the synthetic antipodal grasps, the HOGraspFlow outputs are converted into the same object-centric parallel-jaw representation. Specifically, gripper poses ar transformed into the workspace's coordinate frame, and the corresponding jaw width, approach direction, and contact pair are computed with respect to the object geometry. Grasps that violate the gripper workspace, collide with the object or environment before closure, or fail the validity checks are removed. The remaining antipodal grasps consequently supervise the same contact-location, directional, and grasp quality predictions.

### 3.4 Learning from real-world experiences

The lower-center branch of Fig. 2. corresponds to real-world robot experience. During deployment, the robot executes grasps in cluttered, partially visible, and physically variable environments. Each trial produces observations (e.g., point clouds, RGB images, etc.), predicted grasps, and post-action success, which capture real-world effects that are difficult to model in simulation, including unmodeled contact effects, sensing artifacts, reflective or damaged surfaces, and complex object interactions.

All execution data is added to the shared grasp learning database. This creates a feedback loop in which the system continuously refines its grasp predictions using grounded interaction outcomes. As a result, CIRCA-Grasp does not rely on a fixed offline dataset, but instead incrementally improves its model as new objects, failures, and success cases are encountered during real deployment.

### 3.5 Deployment with active view planning

During deployment, VISO-Grasp uses the current vMF-Contact checkpoint as its grasp inference module. The checkpoint predicts probabilistic contact-grasp candidates and their uncertainty from each acquired observation, while VISO-Grasp determines whether an additional viewpoint is required. Specifically, high uncertainty caused by insufficient observability triggers active view acquisition.

After viewpoint adjustment, the observation is passed to vMF-Contact, which predicts probabilistic grasp candidates together with uncertainty estimates. These estimates play a central role in decision-making. If uncertainty mainly results from limited observability (i.e., epistemic uncertainty), the system seeks a better view. If the scene is sufficiently visible, the model executes the most promising grasp from grasp pools maintained by the Real-Time Uncertainty-guided Multi-view Grasp Fusion (RT-UMGF) module [5]. This coupling of view planning, uncertainty-aware grasp prediction and integration allows the robot to actively search for objects and recover graspable evidence in cluttered environments.

### 3.6 Shared grasp learning database

A key design element of CIRCA-Grasp is the grasp learning database, which unifies data from synthetic generation, human demonstrations, and real robot executions. Each entry may contain scene observations, object hypotheses, candidate grasps, uncertainty estimates, selected viewpoints, and execution labels. This unified representation enables cross-domain learning: simulation contributes coverage and scale, demonstrations contribute semantic and functional priors, and real executions contribute physically grounded feedback. In the proposed adaptation procedure, database records are added after every robot trial, whereas network parameters are not updated inside the real-time control loop. Model adaptation is performed asynchronously at the end of each production shift using source-balanced replay batches containing synthetic grasps, retargeted human priors, and automatically verified robot experiences.

To prevent repetitive and redundant successes from dominating the database, CIRCA-Grasp uses a balanced replay strategy that downsamples redundant low uncertainty successes and prioritizes failures, high uncertainty grasp proposals, and outcomes that disagree with the predicted grasp quality. A representative subset of routine successes is retained to preserve previously learned behavior.

By organizing all sources within the same database, CIRCA-Grasp supports iterative retraining and continual adaptation of the grasp model. Therefore, the database serves as the central memory of the system and the interface through which digital and physical assets are fused.

## 4 Integration into the Knowledge-based Intralogistics

We elucidate the integration of the CIRCA-Grasp framework within a comprehensive knowledge-based intralogistics system designed for circular manufacturing. From the perspective of the knowledge representation of the intralogistics system proposed by [14], CIRCA-Grasp serves as the concrete instantiation of the *:GraspLearningSystem* entity within the factory's overarching knowledge graph.

Fig. 3. explicitly delineates this relationship, mapping the framework's closed-loop architecture to formal ontological classes. At its core, the *central :GraspLearningSystem* governs several interconnected subsystems via hierarchical hasSubSystem relations. The primary data repository is formalized as the *:DataBuffer*, which corresponds directly to the grasp learning database. This buffer systematically aggregates and processes *:Data* from three distinct streams: synthetic digital assets generated in a *:SimulationEnvironment*, human demonstration transfers, and empirical real-world robotic experiences from *:RealWorldEnvironment*. The knowledge graph further formalizes the active learning loop through specialized functional units. The *:TrainingSystem* subsystem encapsulates the core *:NeuralNetworkModel*, specifically the vMF-Contact architecture, which undergoes initialization and training utilizing the aggregated digital assets within the simulation domain. Concurrently, physical deployment is managed by the *:PostProcessingUnit* operating within the *:RealWorldEnvironment*. This unit leverages the predictions from the trained *:NeuralNetworkModel* to be processed by *:postProcessPredictions*, thereby enabling the VISO-Grasp mechanism to execute active view planning and adaptive perception. By feeding these real-world physical executions back into the central *:DataBuffer*, the system establishes a continuous loop of refinement, seamlessly amalgamating scalable synthetic supervision, semantic human priors, and grounded robotic interactions into a unified cognitive architecture.

In summary, CIRCA-Grasp is a systematic and practical instantiation of the knowledge-based concept for the manipulator module from the intralogistics systems, which empowers it to autonomously seek and learn from informative interactions, ensuring rapid policy generalization to unseen and structurally compromised objects.

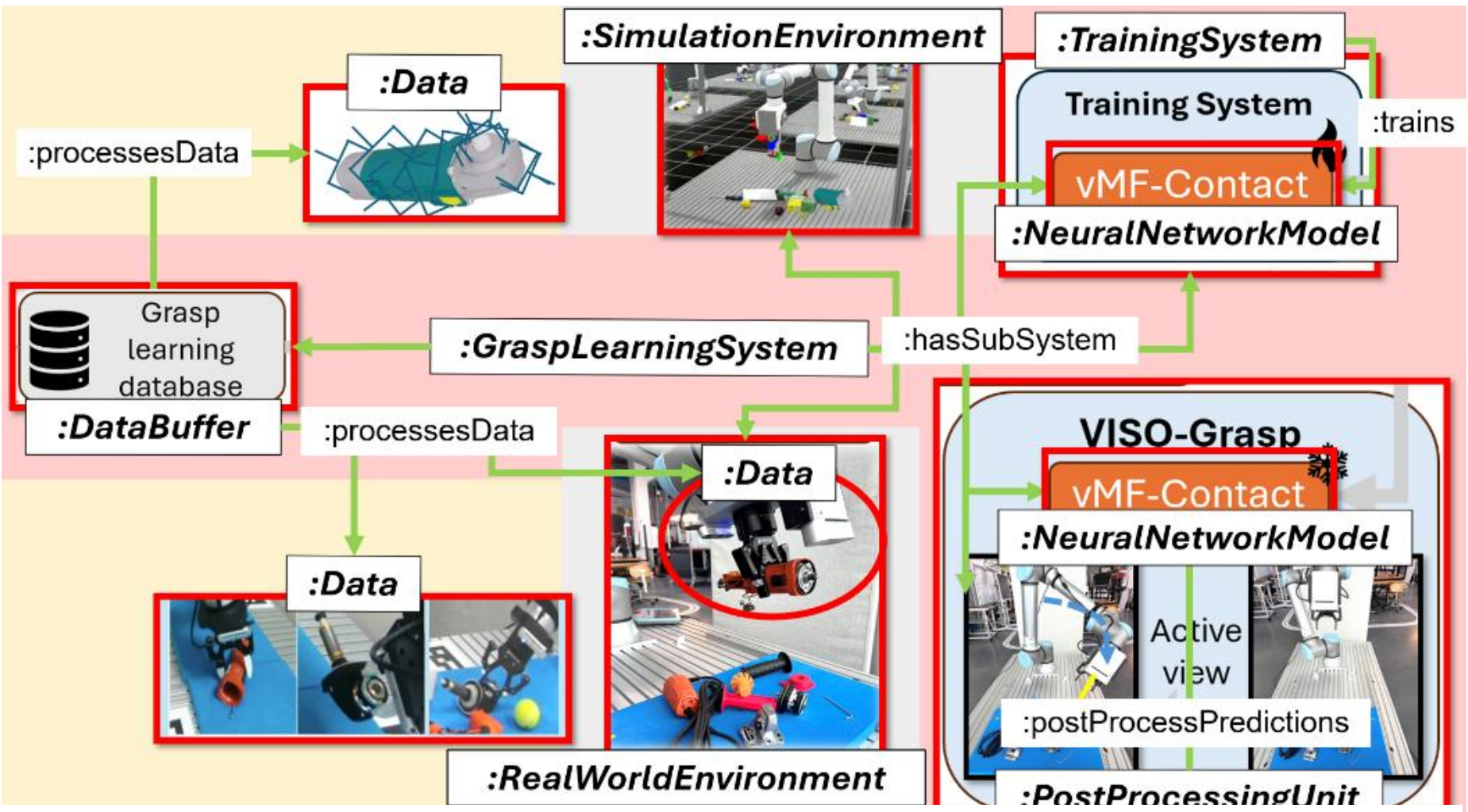

**Fig. 3.** The connection between CIRCA-Grasp and the knowledge-based Intralogistics system from [14]. From the knowledge representation's perspective, CIRCA-Grasp is the instantiation of the *:GraspLearningSystem* within the knowledge graph. (*Note: we use the same color scheme as [14], Fig. 7., to make them comparable)

## 5 Discussion and Limitations

Despite its unified design, CIRCA-Grasp has several limitations. Its computational scalability depends on the cost of probabilistic inference, view planning and execution time, and growth of the experience database. Although offline model adaptation, selective replay, and uncertainty-based termination can reduce these costs, inference latency, memory consumption, and retraining time are expected to be evaluated. Real-time applicability therefore denotes decision latency compatible with the manipulation cycle, although quantitative measurements of inference time, memory use, database growth, and retraining cost remain future work. Besides, the generalizability to embodiments other than parallel end-effectors is limited by the representation of HOGraspFlow retargeting. Extending the framework to suction or multi-fingered grippers would require specific end-effector contact representations, prediction heads, and feasibility checks. Moreover, future work will extend the CIRCA-Grasp framework toward a lifelong learning paradigm [16], enabling incremental adaptation to out-of-distribution shifts and novel component states without catastrophic forgetting.

## 6 Conclusion

We present CIRCA-Grasp, a comprehensive, closed-loop learning framework designed to overcome the profound uncertainties and structural variations inherent in circular manufacturing. By systematically integrating synthetic training data, semantic human interaction priors via HOGraspFlow, and continuous real-world empirical feedback into a shared grasp learning database, the system effectively bridges the gap between structured linear production and dynamic circular environments. Furthermore, our framework's unique coupling of the vMF-Contact model for probabilistic, uncertainty-aware grasp prediction with VISO-Grasp for vision-language-informed active view planning empowers robots to autonomously detect, reveal, and manipulate out-of-distribution and occluded components.

## 7 Acknowledgement

This work is supported by the German Federal Ministry of Research, Technology, and Space (BMFTR) under the Robotics Institute Germany (RIG), the DFG SFB-1574-471687386 project, and the Ministry of Science, Research and Arts of the Federal State of Baden-Württemberg within the InnovationCampus Future Mobility.

## References

1. Newbury, R., Gu, M., Chumbley, L., Mousavian, A., Eppner, C., Leitner, J., Bohg, J., Morales, A., Asfour, T., Kragic, D., et al.: Deep learning approaches to grasp synthesis: A review. IEEE Transactions on Robotics 39(5), 3994–4015 (2023).

2. Gawlikowski, J., Tassi, C.R.N., Ali, M., Lee, J., Humt, M., Feng, J., Kruspe, A., Triebel, R., Jung, P., Roscher, R., et al.: A survey of uncertainty in deep neural networks. Artificial Intelligence Review 56(Suppl 1), 1513–1589 (2023).
3. Shi, Y., Schillinger, P., Gabriel, M., Qualmann, A., Feldman, Z., Ziesche, H., Vien, N.A.: Uncertainty-driven exploration strategies for online grasp learning. In: 2024 IEEE International Conference on Robotics and Automation (ICRA), pp. 781–787. IEEE (2024).
4. Shi, Y., Welte, E., Gilles, M., Rayyes, R.: vmf-contact: Uncertainty-aware evidential learning for probabilistic contact-grasp in noisy clutter. In: 2025 IEEE International Conference on Robotics and Automation (ICRA), pp. 11668–11674. IEEE (2025).
5. Shi, Y., Wen, D., Chen, G., Welte, E., Liu, S., Peng, K., Stiefelhagen, R., Rayyes, R.: VISO-Grasp: vision-language informed spatial object-centric 6-DoF active view planning and grasping in clutter and invisibility. In: 2025 IEEE/RSJ International Conference on Intelligent Robots and Systems (IROS), pp. 14931–14938. IEEE (2025).
6. Sundermeyer, M., Mousavian, A., Triebel, R., Fox, D.: Contact-graspnet: Efficient 6-dof grasp generation in cluttered scenes. In: 2021 IEEE International Conference on Robotics and Automation (ICRA), pp. 13438–13444. IEEE (2021).
7. Gilles, M., Furmans, K. and Rayyes, R.: MetaMVUC: Active Learning for Sample-Efficient Sim-to-Real Domain Adaptation in Robotic Grasping," in IEEE Robotics and Automation Letters, vol. 10, no. 4, pp. 3644-3651. IEEE (2025).
8. Breyer, M., Ott, L., Siegwart, R., Chung, J.J.: Closed-loop next-best-view planning for target-driven grasping. In: 2022 IEEE/RSJ International Conference on Intelligent Robots and Systems (IROS), pp. 1411–1416. IEEE (2022).
9. Pitcher, T., Förster, J., Chung, J.J.: Reinforcement learning for active search and grasp in clutter. In: 2024 IEEE/RSJ International Conference on Intelligent Robots and Systems (IROS), pp. 7301–7306. IEEE (2024).
10. Shi, Y., Guo, Z., Wolf, R., Welte, E., Rayyes, R.: HOGraspFlow: Exploring Vision-based Generative Grasp Synthesis with Hand-Object Priors and Taxonomy Awareness. arXiv preprint arXiv:2509.16871 (2025).
11. Qin, Y., Wu, Y.H., Liu, S., Jiang, H., Yang, R., Fu, Y., Wang, H.: DexMV: Imitation learning for dexterous manipulation from human videos. In: European Conference on Computer Vision (ECCV), pp. 570–587. Springer (2022).
12. Papagiannis, G., Di Palo, N., Vitiello, P., Johns, E.: R+ x: Retrieval and execution from everyday human videos. In: 2025 IEEE International Conference on Robotics and Automation (ICRA), pp. 8284–8290. IEEE (2025).
13. Lepert, M., Fang, J., Bohg, J.: Phantom: Training robots without robots using only human videos. arXiv preprint arXiv:2503.00779 (2025).
14. Klein, J.F., Wolf, R., Ernst, A., Shi, Y., Schumacher, P., Thapa, R.B., Rayyes, R., Furmans, K.: A knowledge-based intralogistic system for a circular factory. Logistics Journal: Proceedings (21) (2025).
15. Mittal, M., Roth, P., Tigue, J., Richard, A., Zhang, O., Du, P., Serrano-Munoz, A., Yao, X., Zurbrügg, R., Rudin, N., et al.: Isaac lab: A gpu-accelerated simulation framework for multimodal robot learning. arXiv preprint arXiv:2511.04831 (2025).
16. Rayyes, R., Donat, H., Steil, J.J.: Hierarchical interest-driven goal babbling for efficient bootstrapping of sensorimotor skills. In: 2020 ICRA, pp. 8303–8309. IEEE (2020). https://doi.org/10.1109/ICRA40945.2020.9196944